\documentclass{article}

\usepackage{microtype}
\usepackage{graphicx}
\usepackage{subfigure}
\usepackage{booktabs} 
\usepackage{hyperref}

\usepackage[accepted]{icml2025}

\usepackage{amsmath}
\usepackage{amssymb}
\usepackage{mathtools}
\usepackage{amsthm}
\usepackage{multirow} 
\usepackage[capitalize,noabbrev]{cleveref}

\theoremstyle{plain}

\theoremstyle{definition}

\theoremstyle{remark}

\usepackage[textsize=tiny]{todonotes}

\icmltitlerunning{EntMTP: Accelerating LLM Inference with Entropy Guided Multi Token Prediction}

\begin{document}




\icmlkeywords{Machine Learning, ICML}

\vskip 0.3in




\twocolumn[
\icmltitle{EntMTP: Accelerating LLM Inference with\\ Entropy Guided Multi Token Prediction}
\begin{center}
\begin{tabular}{cc}
\textbf{Carrie Chen} \\
Cornell University  \\
\texttt{cc2864@cornell.edu} 
\end{tabular}
\end{center}

\vskip 0.3in
]

\vspace{2em}
\begin{abstract}
Multi-token prediction has been shown to increase data density during training, improve downstream text-generation quality, and serves as the defacto approach for self-speculative decoding. Existing foundation and open source models that use MTP heads commit to a static tree-based attention topology throughout the entire generation sequence, meaning the speculation depth, and thus the compute required during verification, stays constant regardless of the context. This is fundamentally misaligned with the entropy patterns of natural language where low-entropy regions often support reliable multi-step drafting, while high-entropy regions require more conservative speculation. To address this, we propose Entropy-guided Multi-Token Prediction (EntMTP), a training-free scheduler that toggles between tree-based attention topologies from a set of task-specific pareto-optimal trees conditioned on a running estimate of local generation entropy. By matching speculation depth to context predictability, EntMTP maximizes expected accepted-token throughput across the full distribution of generated text without sacrificing generation quality. When evaluated across Humaneval, ShareGPT, GSM8k, and Litbench benchmarks, EntMTP consistently achieves a 1.09-1.15x speedup against Hydra and peak speedup of $\sim$1.36x against Medusa baselines respectively. 
\end{abstract}

\section{Introduction}
Large language models (LLMs) have achieved incredible performance across a broad
range of tasks, however the sequential
nature of autoregressive decoding remains a bottleneck to their latency: each generated token requires a full forward pass through all model parameters, making inference memory-bandwidth bound and
expensive at scale.
Speculative decoding~\citep{leviathan2023,chen2023} addresses this by drafting
$k$ candidate tokens with a cheap proposal model and verifying them in a single
parallel pass of the target model, preserving the target distribution exactly.
Multi-token prediction (MTP) heads~\citep{cai2024medusa,ankner2024hydra} implement speculative decoding without a separate draft model by attaching lightweight modules to the target model’s final hidden state and generating a tree of candidates for sparse verification.

For MTP-based decoding, performance is largely determined by the realized \emph{accept depth} at each step. Prior work~\citep{cai2024medusa,ankner2024hydra} uses this metric to select a draft-tree topology offline on a subsample of Alpaca~\citep{wang2023selfinstruct}; the selected topology is then reused throughout inference. We show in Appendix A that this global selection is too coarse. Draft-token acceptance varies with position, local context, and overlap with the training distribution. More importantly, both acceptance behavior and the throughput frontier over tree topologies are \textit{task-specific}: a topology that is Pareto-optimal for one benchmark can be highly suboptimal for another.

We propose EntMTP, a lightweight runtime tree-selection layer that exploits this task dependence rather than treating it as noise. EntMTP first builds a throughput-Pareto frontier of Hydra tree topologies offline, using greedy expansion and self-rollout acceptance estimates. It retains only frontier trees that improve the cost–throughput tradeoff, discarding dominated topologies.

At inference time, a precompiled TopologyBank stores the attention masks, position offsets, and gather indices for each frontier tree. Switching topologies therefore reduces to an $O(1)$ pointer swap, with no mask reconstruction or kernel rebuild. At each decode step, the scheduler reads features off-the-shelf from the model, such as the EAGLE-2 path value, or the base top-1 probability, and selects a tree using one of three policies: \textsc{EntMTP-$l$}, a threshold ladder, \textsc{EntMTP$^*$}, a static-task-optimal tree rule, or \textsc{EntMTP$^\tau$}, a hysteretic binary-$^\tau$ rule with separate on/off thresholds to prevent per-step jitter.

In the $batch\_size =1 $ setting, \textsc{EntMTP$^\tau$} dominates both the Hydra and Medusa baselines as well as \textsc{EntMTP$^*$} across all three evaluation tasks: grade school math, code completion, and conversational llm. On GSM8K it achieves $112.0$ tok/s vs.\ $102.4$ for Hydra
($+9.4\%$), with a speedup of $3.13\times$ over autoregressive decoding. On
HumanEval it achieves $124.7$ tok/s ($+14.0\%$ over Hydra, $3.26\times$ over
AR). The largest absolute  gain is on ShareGPT: $117.5$ tok/s vs.\ $109.0$
($+7.8\%$, $3.47\times$ AR). Response perplexity stays within $0.022$ of
Hydra across all settings, confirming that gains stem from scheduling, not from
changes to the verification rule.
\section{Related Work}

\subsection{Speculative Decoding}
Speculative decoding \citep{leviathan2023, chen2023} accelerates autoregressive generation losslessly by alternating a low-cost \emph{draft} stage with a parallel \emph{verification} stage on the target model.

Given a prefix $T_{1:j}$, a draft distribution $q$ produces $\hat{T}_{j+1:j+k}$ together with its per-token probabilities $\hat{p}_{j+i}$, and the target distribution $p$ then evaluates the entire candidate in a single forward pass.
Tokens are accepted left-to-right with probability $\min(1,\,p_{j+i}(\hat{t}_{j+i})/\hat{p}_{j+i}(\hat{t}_{j+i}))$; on the first rejection, a replacement is drawn from the residual $\mathrm{norm}(\max(0,\,p_{j+i}-\hat{p}_{j+i}))$ and the suffix is discarded.

This ensures that the marginal distribution of accepted tokens matches the target's, so the realized speedup depends entirely on how well $q$ mimics $p$ and how parallelizable verification is.

\subsection{EAGLE and EAGLE-2}

EAGLE~\citep{li2024eagle} replaces token-level drafting with feature-level autoregression. 
Given the target model's hidden-state sequence \(F_{1:i}\) and token sequence \(T_{1:i}\), 
a lightweight drafter predicts the next hidden state,
\[
\hat{F}_{i+1} = D_{\theta}(T_{2:i+1}, F_{1:i})
\]
which is then decoded by the target model's frozen LM head:
$$p_{\theta}(t_{i+1}\mid t_{\le i}) =
\operatorname{softmax}(W_{\mathrm{LM}}\hat{F}_{i+1})$$
The drafter is trained to match the target model's next hidden state using a feature-regression loss,
\[
\mathcal{L}_{\mathrm{reg}}
=
\operatorname{SmoothL1}
\left(
F_{i+1},
D_{\theta}(T_{2:i+1}, F_{1:i})
\right)
\]
By drafting in hidden-state space rather than directly predicting independent future tokens, EAGLE obtains a stronger drafter at comparable parameter cost.

EAGLE-2~\citep{li2024eagle2} keeps this feature-level drafter but makes the draft tree dynamic. 
Instead of committing to a fixed topology, it scores each candidate path using the drafter's own confidence. 
For a node \(t_i\) in the draft tree, its path value is defined as
\[
V_i
=
\prod_{t_j \in \operatorname{Path}(\mathrm{root}, t_i)} p_j
\approx
\prod_{t_j \in \operatorname{Path}(\mathrm{root}, t_i)} c_j
\]
where \(p_j\) is the true, unobserved acceptance probability at node \(t_j\), and \(c_j\) is the drafter's local confidence. 

At inference time, EAGLE-2 expands the tree top-down by repeatedly selecting the highest-value frontier nodes:
\[
t^\star = \arg\max_{t_i \in \mathcal{F}} V_i
\]
where \(\mathcal{F}\) denotes the current frontier. Since the scores \(c_j\) are already produced by the drafter, this dynamic tree construction requires no additional trained components and preserves lossless target-model verification.

\subsection{Independent Multi-Head Prediction}
Medusa~\citep{cai2024medusa} takes a more direct approach to speculation by attaching $K$ lightweight prediction heads to the target model's final hidden state. Each head $h_i$ predicts the token at offset $i$ independently of the other drafted tokens, factorizing the draft distribution as
\[
\prod_i p_{h_i}(t_{j+i}\mid h_j)
\]
The heads emit top-$k_i$ candidates, whose Cartesian product is arranged into a draft tree with precomputed attention masks, position offsets, and gather indices. Verification then reduces to a single sparse forward pass through the target model. 

\subsection{Sequential Multi-Head Prediction}
Hydra~\citep{ankner2024hydra} keeps Medusa's head-on-hidden-state design but restores sequential dependence across draft heads. Instead of predicting each future token independently, head $h_i$ conditions on the embeddings of tokens already drafted by earlier heads, modeling
\[
p(t_{j+i}\mid t_{\le j}, \hat{t}_{j+1:j+i-1})
\]
instead of the marginal $p(t_{j+i}\mid t_{\le j})$

Each head is implemented as a small causal transformer block over the base hidden state and previously drafted token embeddings. This preserves an $O(K)$ head cost while allowing the draft chain to condition on its own prefix. 

However, Hydra inherits Medusa's static-tree limitation: a single topology is selected offline and reused for every decoding step. 

\section{Optimizing Task-Specific Greedy Draft Trees}
\label{sec:per-task-trees}
Hydra's acceptance behavior is workload-dependent in ways the verifier and
base model cannot absorb. On \textsc{ShareGPT}, recent acceptance history
alone predicts the next accept length (EMA $r{=}0.49$); on \textsc{GSM8K}
and \textsc{HumanEval} the strongest predictors are entropy-based and at
least $2{\times}$ weaker ($|r|\!\le\!0.22$) 
(Fig.~\ref{fig:regime-per-task}; full tables in
Appendix~\ref{app:per-task}). 

Since the verifier and base weights are
shared across runs, this dispersion is a property of the workload-induced
token distribution, and implies that single draft topology cannot be optimal across
distributions whose acceptance signal differs by more than $2{\times}$.
We therefore optimize a separate draft tree per task, following the
two-stage offline procedure of \citet{ankner2024hydra,cai2024medusa}.
\subsection{Constructing the Acceptance Frontier}
First, a sequence of proposal trees $T_1, \ldots, T_N$ of increasing size is constructed greedily: 

\begin{algorithm}[H]
\caption{Greedy construction of the acceptance frontier}
\label{alg:greedy-tree}
\small
\begin{algorithmic}[1]
\REQUIRE max depth $D$, budget $B$, prompts $\mathcal{P}$, verifier $\mathcal{V}$
\STATE $T \leftarrow \{(0)\}$, $\mathcal{H} \leftarrow [\,]$
\WHILE{$|T| < B$}
  \STATE $F \leftarrow \textsc{NextChildren}(T, D)$
  \IF{$F = \emptyset$}
    \STATE \textbf{break}
  \ENDIF
  \STATE Score all $T \cup \{\nu\}$ for $\nu \in F$ using one rollout of $T \cup F$
  \STATE $\nu^\star \leftarrow \arg\max_{\nu \in F}
  \bigl(a(\nu), -\sum \nu, -|\nu|\bigr)$
  \STATE $T \leftarrow T \cup \{\nu^\star\}$
  \STATE $\mathcal{H}.\mathrm{append}((|T|, a(\nu^\star), T))$
\ENDWHILE
\STATE \textbf{return} $\mathcal{H}$
\end{algorithmic}
\end{algorithm}

\begin{figure}[ht]
\begin{center}
\centerline{\includegraphics[width=\columnwidth]{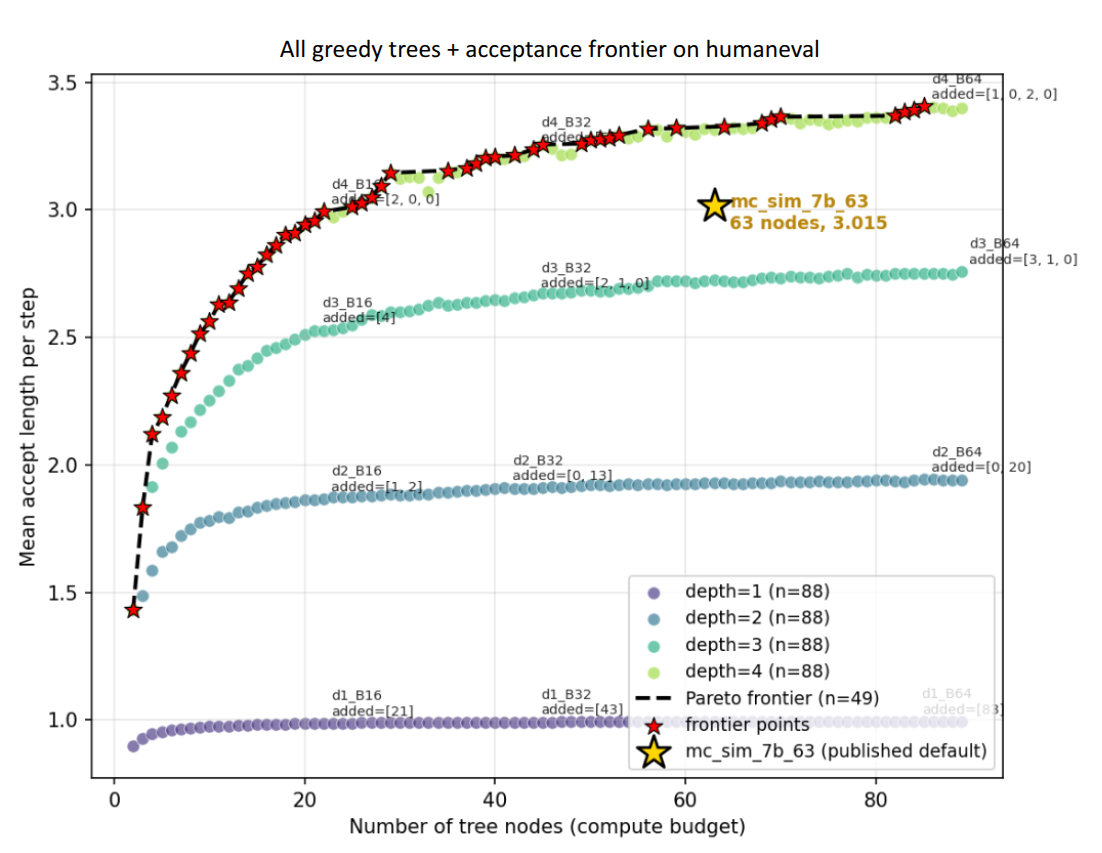}}
\caption{Acceptance frontier on HumanEval. Every candidate tree generated by greedy node addition, colored by depth; the global Pareto hull (red stars, $n{=}49$ (number of items) is dominated by depth-4 trees when nodes $\geq 23$}
\label{fig:acceptance-frontier}
\end{center}
\vspace{-2.5em}
\end{figure}

We score all one-node augmentations $T \cup \{\nu\}$ for $\nu \in F(T)$ in a single self-rollout over the joint tree $T \cup F(T)$, recovering each candidate’s accept length by masking verifier logits along its path.\footnote{The trajectory is advanced using $T$, not the augmented scoring tree, so this is a one-step ``what-if'' probe.} The frontier node with the highest mean accept length is added, breaking ties by smaller child-rank sum and shallower depth, and repeat until the target budget is reached (Fig.~\ref{fig:acceptance-frontier}). 

All runs follow the same format of $100$ prompts from the target benchmark, $T{=}0.7$, posterior threshold $0.09$, $\alpha{=}0.3$, $256$-token rollouts.

\subsubsection{One-Step Scoring Bias}
The mean-accept estimate computed inside Algorithm~\ref{alg:greedy-tree}
is mildly optimistic. Candidate $T \cup \{\nu\}$ is scored against
proposals drawn from the larger joint tree $T \cup F$, so the verifier
sees draft tokens at positions $\nu$ would not have proposed on its own. Before the second stage, frontiers are
recalibrated against a canonical Hydra self-rollout that uses each candidate tree's own proposal, attention mask, and KV-cache update
(Appendix~\ref{app:debias}).

\subsection{Constructing the Throughput Frontier}
Tree shape also sets the
per-step verifier cost through the size of the attention mask, the number of
candidate paths returned by \textsc{tree\_decoding}, and the cost of
\textsc{evaluate\_posterior}~\citep{ankner2024hydra}. The second stage evaluates the entire per-depth acceptance frontier end-to-end on the target workload, recording wall-clock output tokens per second under the same
proposal $\rightarrow$ tree decoding $\rightarrow$ verification $\rightarrow$
KV-cache update loop used at inference time, including prefill. Selecting
the throughput-maximal tree on this re-ranked hull gives the task-specific
tree topology used in \textsc{EntMTP$^*$}.

\begin{figure}[ht]
\vspace{-1em}

\begin{center}
\includegraphics[width=0.95\columnwidth]{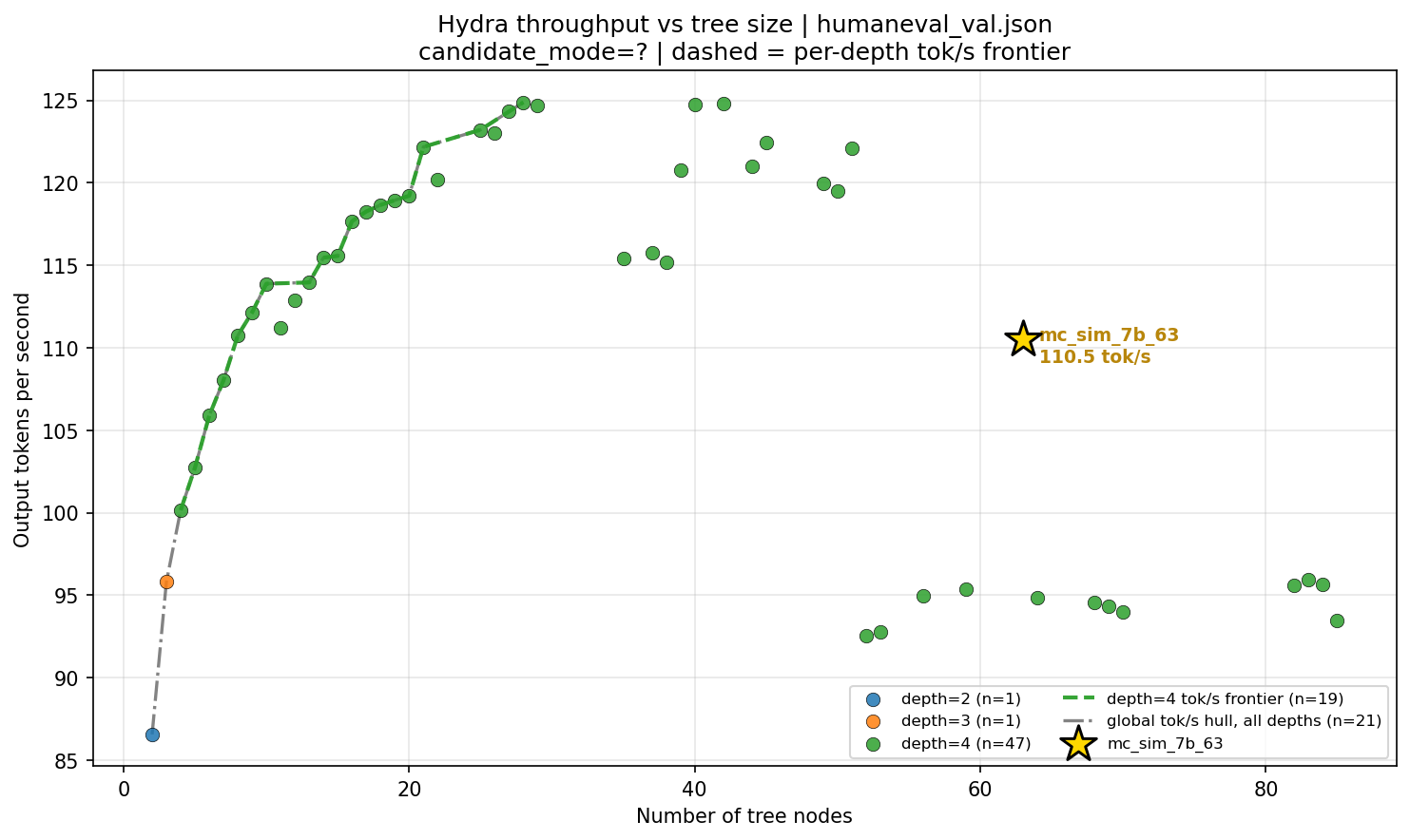}
\caption{Re-ranking HumanEval's acceptance
frontier ($49$ candidate trees plus the published default) over $100$ HumanEval-val prompts.}
\label{icml-historical}
\end{center}
\vspace{-2em}
\end{figure}

\section{Inference-Time Tree Scheduler}
Section~\ref{sec:per-task-trees} produces, for each benchmark, a small
throughput-Pareto bank of draft trees $\mathcal{B} = \{T_1, \dots, T_K\}$.
Selecting any fixed $T_k$ already beats the published default
(Section \S\ref{sec:results}), but the regime analysis in
Fig.~\ref{fig:regime-per-task} shows the optimal tree size is itself
\emph{step-dependent}: long high-acceptance runs reward a large tree,
while regions of high entropy or recent rejections reward a cheap one.
We therefore add a per-step policy $\pi : \mathbb{R}^d \!\to\! \{1,\ldots,K\}$
that maps the current decode state to an index into $\mathcal{B}$.
\paragraph{Path-value score.}
The features the policy consumes are the EAGLE-2 path values
\citep{li2024eagle2}, computed at the verifier's last position from the
base-LM top-1 probability $p_0$ and each Hydra head's top-1 probability
$p_d$. The depth-$d$ greedy path probability is
$g_d = p_0 \prod_{i=1}^{d} p_i$, and the policy uses the scalar
$s = \max(p_0, g_1, \ldots, g_D) = \texttt{best\_path\_value}$,
which is monotone in the verifier's expected accept length: when the head
chain has high mass deep, $s \!\to\! 1$ and a large tree is worth its cost.
\paragraph{Policy.}
We instantiate $\pi$ as a binary thresholded selector with hysteresis. Two
trees from $\mathcal{B}$ are designated: a conservative $T_-$ (small,
shallow) and an aggressive $T_+$ (large, depth-$4$). At each step,
$\pi$ switches to $T_+$ when $s > \tau_{\text{on}}$ and back to $T_-$
when $s \le \tau_{\text{off}}$, with $\tau_{\text{off}} \le \tau_{\text{on}}$
to suppress flapping. A single threshold ($\tau_{\text{on}}\!=\!\tau_{\text{off}}\!=\!\tau$) is the default, and the
$K$-way generalisation \texttt{threshold\_ladder} bins $s$ into $K$
consecutive intervals selecting $T_1, \ldots, T_K$. Both policies are
training-free: $^\tau$ is picked by a one-dimensional sweep on the same
$100$-prompt calibration set used for tree search.
\paragraph{Cost.}
Inserting $\pi$ into the decode loop is essentially free. Each tree in
$\mathcal{B}$ ships with its own precomputed
\texttt{generate\_hydra\_buffers} output (attention mask, position ids,
retrieve indices), so switching trees is a dictionary pointer swap with no
GPU work. Computing $s$ from the verifier logits already in registers
takes one cumulative product over $D{\le}4$ scalars and a single
\texttt{.item()} sync, adding ${<}0.1\,\mathrm{ms}$ per step.
\section{Evaluation Methodology}
\label{sec:eval}
\subsection{Metrics}
\label{sec:metrics}
EntMTP neither fine-tunes the original LLM nor relaxes Hydra's typical
acceptance condition, so it is a lossless acceleration method
(continuation perplexity matches the base LM to within $0.02$ nats in
all our runs). We therefore evaluate acceleration only, with two metrics:
\begin{itemize}
  \item \textbf{Speedup ratio} $\rho$: wall-clock output tokens per second
        relative to vanilla autoregressive decoding of the same base LM
        on the same prompt basket, including prompt prefill in the timed
        region.
  \item \textbf{Average acceptance length} $^\tau$: expected number of
        tokens generated per draft-verify cycle. $^\tau$ is independent
        of hardware and runtime and isolates the quality of the draft
        topology from kernel-level effects.
\end{itemize}
\subsection{Setup}
\label{sec:setup}
We use Vicuna-7B v$1.3$ \citep{vicuna2023} as the base LM and
\texttt{ankner/hydra-vicuna-7b-v1.3} \citep{ankner2024hydra} as the Hydra
verifier, in FP16 on a single NVIDIA A100. All methods share temperature
$T{=}0.7$, posterior threshold $\epsilon{=}0.09$, mixing coefficient
$\alpha{=}0.3$, max input $1400$ tokens, and max generation $256$
tokens. Each benchmark is timed on $100$ prompts (seed $123$) drawn from
HumanEval-val, GSM8K-val, and ShareGPT (Vicuna unfiltered split). Each
run begins with one warm-up generation to amortise JIT and KV-cache
allocation; timings exclude this warm-up. The conservative and
aggressive trees for the scheduler are chosen from the per-task
throughput frontier; $^\tau$ is selected from
$\{0.001, 0.005, 0.01, 0.02, 0.05\}$ on the calibration set.

\section{Results}
\label{sec:results}
The following reports speedup and average acceptance
length of both \textsc{EntMTP} policies across all three benchmarks.
\begin{table}[ht]
\label{tab:main-results}
\vskip 0.05in
\centering
\small
\setlength{\tabcolsep}{4pt}
\begin{tabular}{llrrr}
\toprule
benchmark & method & tok/s & $\rho$ & $\tau$ \\
\midrule
\multirow{4}{*}{HumanEval}
  & Vanilla Vicuna       & $38.2$  & $1.00\!\times$ & $1.00$ \\
    & Medusa (default)       & $91.2$  & $2.38\!\times$ & $2.87$ \\
  & Hydra (default)      & $109.0$  & $2.85\!\times$ & $3.06$ \\
  & \textsc{EntMTP$^*$} & $123.4$ & ${3.21\!\times}$ & $\mathbf{3.28}$ \\
  & \textsc{EntMTP$^\tau$}  & $124.7$ & $\mathbf{3.26\!\times}$ & $3.20$ \\
\midrule
\multirow{4}{*}{GSM8K}
  & Vanilla Vicuna       & $33.7$  & $1.00\!\times$ & $1.00$ \\
& Medusa (default)       & $87.4$  & $2.59\!\times$ & $2.51$ \\
  & Hydra (default)      & $102.4$ & $2.87\!\times$ & $2.86$ \\
  & \textsc{EntMTP$^*$} & $109.7$ & $3.07\!\times$ & $\mathbf{3.08}$ \\
  & \textsc{EntMTP$^\tau$}  & $112.0$ & $\mathbf{3.13\!\times}$ & $3.02$ \\
\midrule
\multirow{3}{*}{ShareGPT}
  & Vanilla Vicuna       & $35.4$  & $1.00\!\times$ & $1.00$ \\
  & Medusa (default)       & $94.6$  & $2.72\!\times$ & $2.86$ \\
   & Hydra (default)      & $109.0$  & $2.89\!\times$ & $3.06$ \\
  & \textsc{EntMTP$^*$} &  $116.9$ & $3.42\!\times$ & $2.97$ \\
  & \textsc{EntMTP$^\tau$}  & $117.5$ & $\mathbf{3.47\!\times}$ & $\mathbf{2.99}$ \\
\bottomrule
\end{tabular}
\caption{Speedup $\rho$ over vanilla Vicuna-7B and average acceptance
length $^\tau$ on $100$-prompt baskets at $T{=}0.7$. Hydra and Medusa use
their authors' published default trees. \textsc{EntMTP}$^{*}$ uses the
per-task throughput-optimal tree; \textsc{EntMTP}$^\tau$ switches per
step between a conservative and an aggressive tree from the same
frontier. All times include prompt prefill.}
\vskip -0.1in
\end{table} 

The fixed per-task throughput-optimal tree
(\textsc{EntMTP}$^{*}$) outperforms Hydra's published default by
$7.1$-$13.2\%$ in tokens/s while using ${\ge}\,2{\times}$ fewer draft
nodes ($28$/$46$/$30$ vs.\ $63$ on HumanEval/GSM8K/ShareGPT), and
outperforms Medusa's default by $7.7$-$32.4\%$. Adding the runtime
scheduler (\textsc{EntMTP}$^\tau$) gives a further $0.5$-$2.1\%$ on top
of \textsc{EntMTP}$^{*}$ and yields the best wall-clock throughput on
every benchmark, $3.26{\times}$ over vanilla Vicuna on HumanEval,
$3.13{\times}$ on GSM8K, and $3.47{\times}$ on ShareGPT. Continuation
perplexity stays within $0.02$ nats of the base LM on every row, so the
gains are lossless.

\subsection{Where did the gains come from?}
The static gain over Hydra's default decomposes consistently across
workloads: most of the $7$-$13\%$ comes from a smaller per-task tree
shrinking per-step verifier cost, with $0$-$7\%$ added by a higher
$^\tau$ on the optimized topology (HumanEval gains both, $^\tau$
$3.06\!\to\!3.28$; ShareGPT trades $^\tau$ down by $3\%$ for a smaller
tree and still nets $+7\%$ tok/s). The schedulers residual gain over
the static tree is largest on GSM8K ($+2.1\%$), where mixing in the
conservative tree during high-entropy reasoning steps reclaims verifier
cycles without losing accept length.

\section{Acknowledgements}
We'd like to thank Professor De Sa for all the amazing conversations and support throughout the semester for this project!!

\nocite{langley00}

\bibliography{example_paper}
\bibliographystyle{icml2025}

\newpage
\appendix
\onecolumn
\label{app:per-task}

\section{Entropy regimes across benchmarks}
\begin{figure}[ht]
\begin{center}
\label{features}
\centerline{\includegraphics[width=\columnwidth]{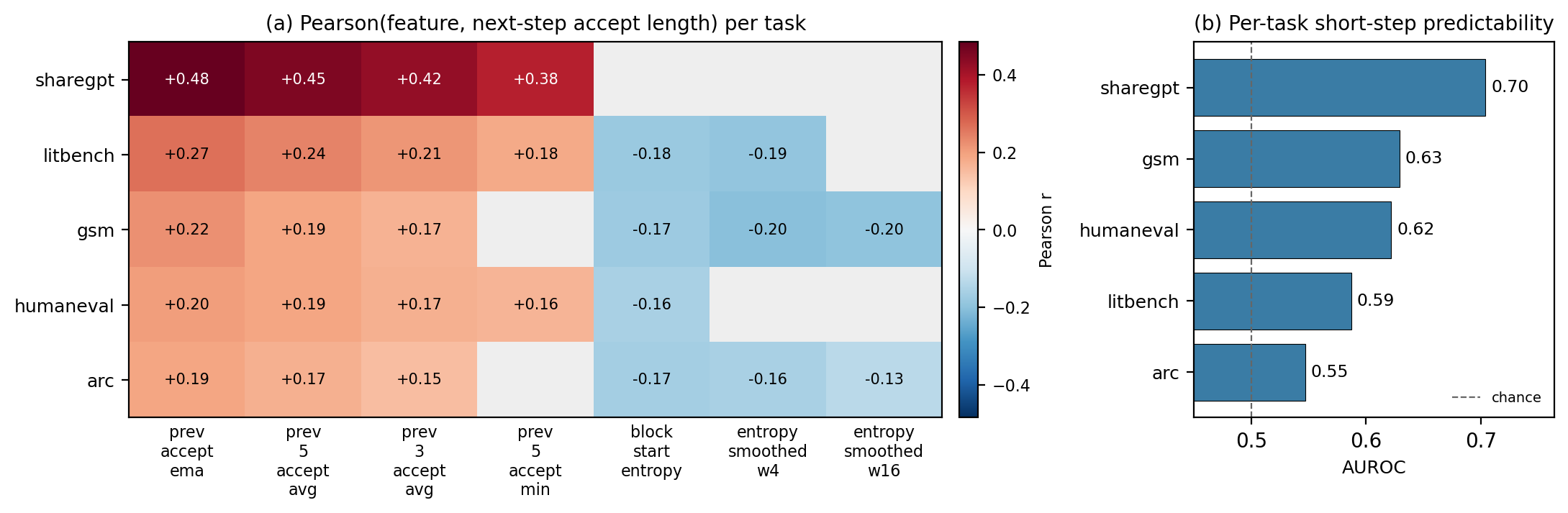}}
\caption{Per-task divergence of speculative-decoding regimes on Vicuna-7B
(see Appendix~\ref{app:per-task} for the full feature lists and model
configurations). \textbf{(a)} Pearson correlation between curated features
and the next-step accept length; gray cells did not enter the per-task top-7.
History features (left) dominate on \textsc{ShareGPT}/\textsc{LitBench};
entropy features (right) are the strongest predictors on
\textsc{GSM8K}/\textsc{HumanEval}/\textsc{ARC}. \textbf{(b)} AUROC of an MLP
predicting whether the next step accepts only one token. Predictability spans
$0.55$-$0.70$ across tasks, motivating per-task tree selection.}
\label{fig:regime-per-task}
\end{center}
\vspace{-1.5em}
\end{figure}

For each benchmark, we log the following at every Hydra decode step in Fig.~\ref{fig:regime-per-task}: block-start
entropy with expoential smoothing ($w \in \{4, 8, 12, 16, 24, 32, 64\}$),
recent-acceptance moments (EMA, last-$k$ mean/min/max), Hydra-head
agreement, confidence statistics, and base-LM token-shape signals
producing $13$k-$320$k step rows per task (HumanEval $13$k, ARC $12$k,
GSM8K $119$k, ShareGPT $160$k, LitBench $320$k). Pearson-$r$ in
Fig.~\ref{fig:regime-per-task}(a) is computed against the next step's
accept length within the same generation trajectory, and the AUROC bars
in (b) come from a small per-task MLP ($64\!\to\!32\!\to\!1$, ReLU +
dropout $0.4$, early-stopped on AUROC) trained to predict
$\mathbb{1}[\textrm{accept\_length}_{t+1}{=}1]$. Splits are taken over
\emph{generation samples} (\texttt{sample\_idx}), not over rows, so
adjacent decode steps from the same prompt never appear on both sides of
the train/test boundary.

\section{Debiasing greedy acceptance frontier}
\label{app:debias}
To ensure that acceptance rates reflect the actual test-time model performance, 
we re-evaluate every candidate tree topology on the per-depth frontier with a
canonical self-rollout that uses the candidate tree's own proposal,
attention mask, and KV-cache update to produce unbiased acceptance estimate. Across all four evaluated tasks 
the correction is small and rank-preserving: the biased and unbiased
means differ by at most $3.3\%$ on any frontier tree, and the resulting
frontier topology is unchanged.

\begin{table*}[h]

\label{tab:debia}
\vskip 0.05in
\centering
\small
\setlength{\tabcolsep}{4pt}
\begin{tabular}{rrrrr@{\hskip 1.8em}rrrrr@{\hskip 1.8em}rrrrr}
\toprule
\multicolumn{5}{c}{\textbf{HumanEval}} & \multicolumn{5}{c}{\textbf{GSM8K}} & \multicolumn{5}{c}{\textbf{ShareGPT}} \\
\cmidrule(r){1-5} \cmidrule(lr){6-10} \cmidrule(l){11-15}
$n$ & $d$ & biased & self & $\Delta$ & $n$ & $d$ & biased & self & $\Delta$ & $n$ & $d$ & biased & self & $\Delta$ \\
\midrule
 2 & 2 & 1.431 & 1.403 & $+2.0\%$ &  2 & 2 & 1.384 & 1.370 & $+1.0\%$ &  2 & 2 & 1.427 & 1.383 & $+3.2\%$ \\
14 & 4 & 2.750 & 2.783 & $-1.2\%$ & 11 & 4 & 2.593 & 2.650 & $-2.1\%$ & 12 & 4 & 2.619 & 2.548 & $+2.8\%$ \\
28 & 4 & 3.095 & 3.119 & $-0.8\%$ & 23 & 4 & 2.888 & 2.841 & $+1.6\%$ & 27 & 4 & 3.009 & 2.949 & $+2.0\%$ \\
51 & 4 & 3.279 & 3.266 & $+0.4\%$ & 41 & 4 & 3.099 & 3.109 & $-0.3\%$ & 52 & 4 & 3.228 & 3.186 & $+1.3\%$ \\
85 & 4 & 3.407 & 3.382 & $+0.7\%$ & 88 & 4 & 3.291 & 3.278 & $+0.4\%$ & 76 & 4 & 3.396 & 3.332 & $+1.9\%$ \\
\bottomrule
\end{tabular}
\caption{Biased (Algorithm~\ref{alg:greedy-tree}, batched scoring) vs.\
unbiased (canonical self-rollout)
mean accept length per step, on five frontier trees per benchmark
spanning the node-budget range. $\Delta$ is the relative bias of the
batched estimator $(\textrm{biased}-\textrm{self})/\textrm{self}$.
Across all $124$ evaluated frontier trees the bias stays within
$\pm 3.3\%$, and the resulting frontier topology is unchanged on every
benchmark.}
\vskip -0.1in
\end{table*}
\section{More Frontiers}
\begin{figure}[t]
\centering
\includegraphics[width=\columnwidth]{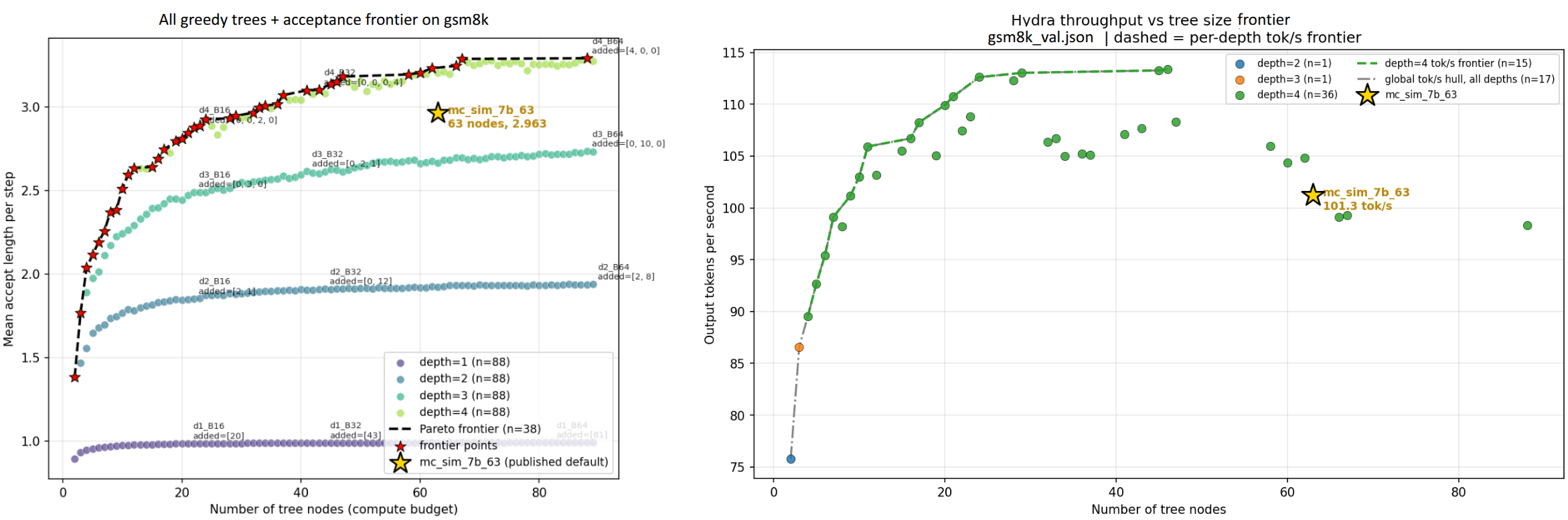}
\caption{Two-stage greedy draft tree frontiers on GSM8K, a multi-step reasoning benchmark for grade school mathematics. \textbf{(a)} Acceptance-rate Pareto frontier, Medusa's published default is pareto-dominated by all depth-4 trees after nodes $\geq 34$; \textbf{(b)} the acceptance frontier (red) re-evaluated on throughput (tokens/second).}
\label{fig:gsm8k-frontier}
\end{figure}

\begin{figure}[t]
\centering
\includegraphics[width=\columnwidth]{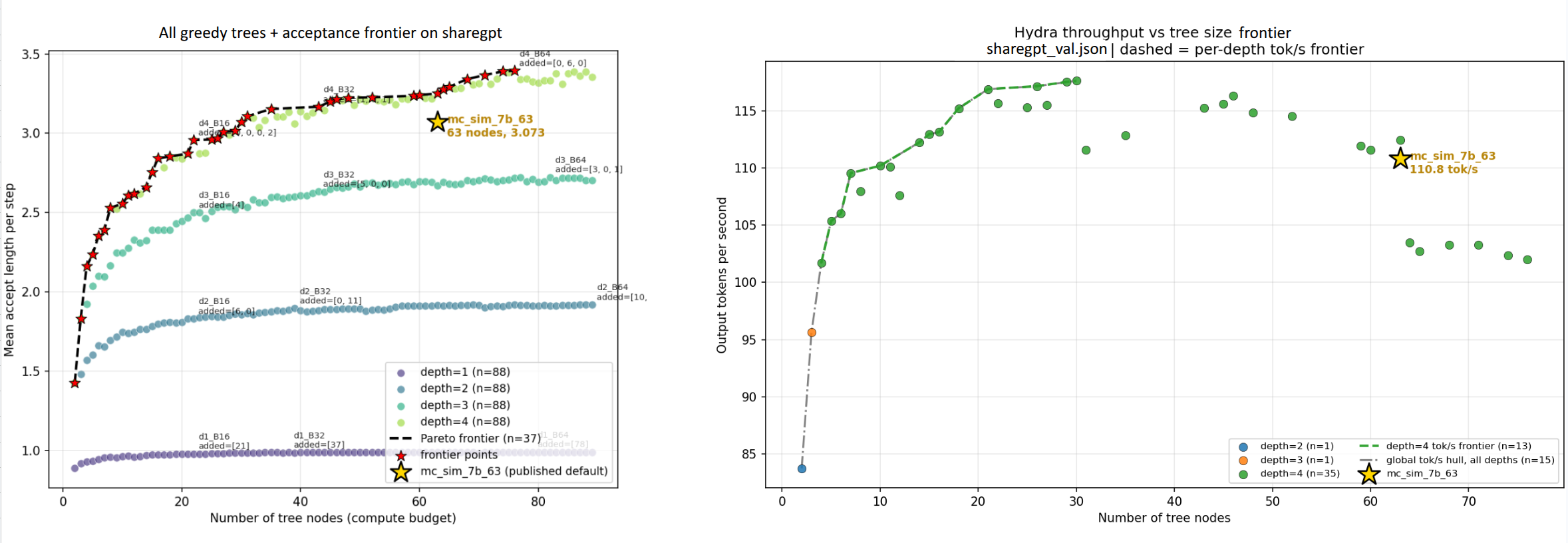}
\caption{Two-stage greedy draft tree frontiers on ShareGPT, a collection of real-world multi-round conversations between users and ChatGPT. \textbf{(a)} Acceptance-rate Pareto frontier, the published default is once again pareto-dominated on task-calibrated depth-4 trees with nodes $\geq 27$; \textbf{(b)} the acceptance frontier mapped onto throughput space.}
\label{fig:sharegpt-frontier}
\vspace{-1.5em}
\end{figure}

Fig.~\ref{fig:gsm8k-frontier} and~\ref{fig:sharegpt-frontier} report
the same two-stage tree search of \S\ref{sec:per-task-trees} applied to
GSM8K and ShareGPT, with calibration and timing protocols identical to
the HumanEval frontier in Fig ~\ref{alg:greedy-tree} and ~\ref{icml-historical}.

$ $
\end{document}